\documentclass[11pt,a4paper]{article}
\usepackage[hyperref]{acl2021}
\usepackage{times}
\usepackage{latexsym}

\usepackage{microtype}
\usepackage{xcolor}
\usepackage{amsmath}
\usepackage{amsfonts}
\usepackage{graphicx}
\usepackage{algorithm,algpseudocode}
\usepackage{algorithmicx}
\usepackage{multirow}
\usepackage{bm}

\aclfinalcopy 

\makeatletter
\newcommand{\thickhline}{
\noalign {\ifnum 0=`}\fi \hrule height 1.0pt
\futurelet \reserved@a \@xhline
}
\makeatother

\algnewcommand\LeftComment[2]{%
\hspace{#1\algindent}$\triangleright$ \eqparbox{}{#2} \hfill %
}

\title{Adaptive Sparse Transformer for Multilingual Translation}

\author{Hongyu Gong, Xian Li, Dmitriy Genzel \\
  Meta AI \\
  \texttt{\{hygong,xianl,dgenzel\}@fb.com} \\}

\date{}

\begin{document}
\maketitle
\begin{abstract}
Multilingual machine translation has attracted much attention recently due to its support of knowledge transfer among languages and the low cost of training and deployment compared with numerous bilingual models. A known challenge of multilingual models is the negative language interference. In order to enhance the translation quality, deeper and wider architectures are applied to multilingual modeling for larger model capacity, which suffers from the increased inference cost at the same time. It has been pointed out in recent studies that parameters shared among languages are the cause of interference while they may also enable positive transfer. Based on these insights, we propose an adaptive and sparse architecture for multilingual modeling, and train the model to learn shared and language-specific parameters to improve the positive transfer and mitigate the interference. The sparse architecture only activates a sub-network which preserves inference efficiency, and the adaptive design selects different sub-networks based on the input languages. Our model outperforms strong baselines across multiple benchmarks. On the large-scale OPUS dataset with $100$ languages, we achieve $+2.1$, $+1.3$ and $+6.2$ BLEU improvements in one-to-many, many-to-one and zero-shot tasks respectively compared to standard Transformer without increasing the inference cost.
\end{abstract}

\section{Introduction}

Multilingual neural machine translation (MNMT) develops one  model for translations in multiple language directions \cite{tan2019multilingual}. A key advantage of multilingual models is the knowledge transfer, which improves the translation performance especially for low-resource languages \cite{zoph2016transfer}. Multilingual models tend to generalize better compared with bilingual translation due to the exposure to diverse languages  \cite{zoph2016multi,arivazhagan2019massively}. Moreover, it is burdensome to train hundreds of bilingual models for each language pair, and one multilingual model reduces the deployment and maintenance cost \cite{dabre2020survey}.

A known challenge for multilingual modeling is the curse of multilinguality, where the language interference hurts model performance \cite{conneau2019unsupervised}.  
Language adapters attract research attention due to their strong performance in cross-lingual modeling \cite{wang2019compact}. Adapter layers are added for each language or language direction, preserving the knowledge of language specificity in a multilingual model \cite{bapna2019simple,zhang2021share}. Despite the simplicity, adapters are faced with extra inference cost brought by additional adapter layers. 

It has been revealed that interference occurs in the shared parameters \cite{wang2020negative}. 
Sparse modeling is extensively studied as an efficient approach to language interference mitigation. With the assumption that interference is more likely between diverse languages, existing works group languages into families based on their proximity \cite{tan2019multilingual}. Languages in the same family are sharing more parameters to encourage positive transfer, and different families have their exclusive decoders \cite{sen2019multilingual}. But as pointed out by \cite{lin2019choosing}, the factors affecting parameter sharing are more complicated than the language proximity. 
 
Recent studies integrate sparsity into different components of a multilingual model and learn to share parameters more flexibly. Latent depth model leverages sparsity across layers, allowing different languages to choose a subset of layers in a deep Transformer model \cite{li2020deep}. GShard \cite{lepikhin2020gshard} and Switch Transformer \cite{fedus2021switch} explore feed-forward (FFN) sparsity with Mixture-of-experts (MoE). They replace a feed-forward sub-layer with a set of identical sub-layers (i.e., multiple experts), and route input tokens to different experts. Language-sensitive attention sparsity is studied and each language direction is assigned with one attention module \cite{wang2019compact}. A more fine-grained scheme of attention sparsity assigns each language with a selected subset of attention heads \cite{gong2021pay}. However, it remains unknown which type of sparsity is most effective in multilingual translation.


In this work, we propose a latent variable model to leverage the language-dependent sparsity in multilingual machine translation.
The latent variables learn to activate various sub-networks for given languages in order to optimize translations.
This is a general approach to integrate sparsity at different scales. We are able to bring different types of sparsity under a common umbrella, including the sparsity within feed-forward and attention module as well as the sparsity across Transformer layers. Therefore, it enables a direct comparison of the sparsity in different components. Moreover, our design supports adaptive sparsity so that we could specify the amount of model sparsity and easily control the inference cost.

One limitation of existing approaches is the coarse-grained parameter sharing, where two languages share either all or no parameters of a given component (e.g., the whole FFN or attention module). It is possible that positive transfer occurs in some parameters while negative interference in other parameters of a component. We propose more fine-grained strategies of partial parameter sharing. For FFN sparsity, our model divides the FFN weight matrix into multiple blocks. Each language activates a subset of blocks so that partial sharing of FFN is enabled among languages.
As for attention sparsity, our model learns to select a subset of attention heads for each language. Knowledge transfer is enabled in shared heads and language specificity is preserved in heads exclusively owned by some languages.



Our main findings are summarized below: 
\begin{enumerate}
    \item Our sparse model consistently outperforms strong multilingual baselines across benchmark datasets on multilingual translation. On the large-scale OPUS dataset with $100$ languages, we achieve average gains of $2.1$, $1.3$ and $6.2$ BLEU over Transformer in one-to-many, many-to-one and zero-shot translations respectively.
    \item Our model preserves the inference efficiency with adaptive sparsity. By controlling the amount of model sparsity, we improve the translation quality without increasing the inference cost.
    \item We compare the sparsity in different model components. FFN and attention sparsity works best for one-to-many and many-to-translation respectively on medium-scale translation with $24$ languages. As for large-scale translation covering $100$ languages, combining all types of sparsity yields the optimal performance.
    \item We analyze the sparsity patterns learned for languages, and reveal that parameter sharing is affected by language proximity as well as the resource sizes.
\end{enumerate}

\section{Related Works}


\begin{figure*}[htbp!]
\centering
\includegraphics[width=\textwidth]{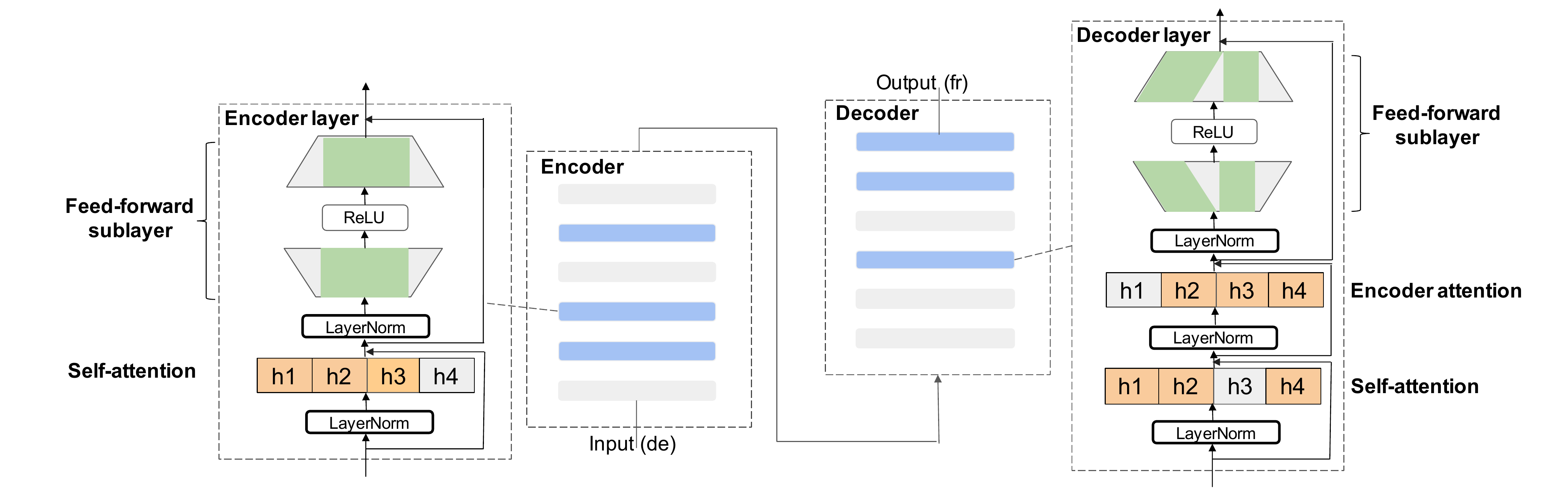}
\caption{Sparse architecture of adaptive Transformer. The grey areas are where sparsity is added to Transformer layers, attention and feed-forward sub-layers. Only a sub-network is activated based on the source and target languages during inference time. In this example, blue layers, orange attention heads and green feed-forward blocks are activated components for source language (de) and target language (fr).  
}
\label{fig:model}
\end{figure*}

\textbf{Multilingual translation}. 
Multilingual translation model refers to a universal system capable of translating between multiple language pairs. 
Multilingual models are appealing due to better scalability, lower maintenance cost and more knowledge transfer compared with bilingual models \cite{dabre2020survey}.
Despite the benefits above, multilingual models are faced with the negative transfer brought by the language interference \cite{conneau2019unsupervised}. 
Recent studies point out that conflicting gradients of different languages in their shared parameters are the cause of negative transfer among languages \cite{yu2020gradient}.

There are two lines of research towards the mitigation of language interference. One line of studies resolve the gradient conflicts during model training from the optimization perspective \cite{suteu2019regularizing,yu2020gradient,wang2020gradient}. The other line of research focuses on the model architecture, and explores ways of parameter sharing \cite{sachan2018parameter}. Our study falls into the category of architecture design. We introduce commonly used approaches in this category such as sparse models and language adapters.

\noindent\textbf{Sparse modeling and conditional sparsity}. Sparse models have been extensively studied, and the lottery ticket hypothesis suggests that sparsity improves efficiency over dense models without hurting model performance \cite{frankle2018lottery}. 
Attention sparsity is introduced to Transformer, where attention weights \cite{correia2019adaptively} or attention outputs \cite{michel2019sixteen} are sparsified. 
However, these approaches are limited to bilingual settings.

Conditional sparsity is further proposed for multilingual models so that the model sparsity is conditioned on languages or input tokens. An early approach trains multiple models for multilingual translation \cite{tan2019multilingual}. Models are independently trained for each language family.
Multi-decoder model routes languages in different families to different decoders \cite{sen2019multilingual,xiang2021multi}.  The intuition behind language grouping is that similar languages tend to have positive transfer. One weakness of having separate modules per language is the increase of model sizes with the number of languages. External linguistic knowledge and expertise are also required to measure the language proximity. Moreover, it has been revealed that high similarity between languages does not always lead to positive transfer \cite{lin2019choosing}.

GShard \cite{lepikhin2020gshard} and Switch Transformer \cite{fedus2021switch} replaces a single FFN sub-layer with Mixture-of-Experts consisting of multiple FFN sub-layers. Each incoming token is routed to one of these FFN experts. 
Language-sensitive attention is integrated into a multilingual model and each language direction has its own cross-attention module \cite{wang2019compact}. It has the limitation of being unable to deal with zero-shot translation. A recent work selects a subset of attention heads for each language, and demonstrates improvements in both multilingual and multi-domain modeling \cite{gong2021pay}.
Conditional sparsity has also been explored across layers, where a deep Transformer is trained to allow different languages to select their own subset of layers \cite{li2020deep}. 

\noindent\textbf{Language adapters}.  \citeauthor{bapna2019simple} inserts an adapter layer for each language pair into a  multilingual Transformer pre-trained in all languages \cite{bapna2019simple}. The adapted model is finetuned separately for each language pair. A similar work uses adapter layers based on languages instead of language directions \cite{philip2020language}. \citeauthor{zhang2021share} adds an adapter layer on top of each attention and FFN sub-layer, and the inputs could choose to used shared or language-specific parameters within the adapter layer \cite{zhang2021share}. The limitations of language adapters include the increasing memory consumption with the number of languages and extra computation costs of additional adapter layers.

\section{Model}

In this work, we propose a sparse multilingual Transformer in order to optimize the translation quality with controlled inference cost. A Transformer model is basically a stack of Transformer layers which consist of multi-head attention and feed-forward sub-layers. We add sparsity to these components respectively. To encourage positive transfer and mitigate negative interference among different languages, the sparsity is language dependent, i.e., the model activates different components for each language. We start with the architecture of a standard Transformer layer, and then introduce sparsity to different modules in a Transformer model.


\subsection{Transformer Layer}
Within a Transformer layer, suppose that there are $H$ heads in its attention module. Each attention head keeps a set of query, key and value vectors for input tokens.
For a given token, a head assigns its attention to the input sequence using query-key matching between tokens. The value vectors of all tokens are weighted by the attention, and the weighted vectors from different heads are concatenated as the new representation of the target token. Let $\mathbf{x}_{h}$ be the output of head $h$, and the token vector $\mathbf{x}$ learned by the attention module is:
\begin{align}
    \mathbf{x} = \mathbf{x}_{1} \oplus \cdots \oplus \mathbf{x}_{h} \oplus \cdots \oplus \mathbf{x}_{H},
\end{align}
where $\oplus$ is the vector concatenation.

The token vector $\mathbf{x}$ from the attention module is updated to $\mathbf{y}$ after linear projection, residual connection and layer normalization. The feed-forward module processes $\mathbf{y}$ with two dense layers. Tokens are projected to a higher dimension space in the first FFN sub-layer and then transformed to the original dimension in the second FFN sub-layer.
\begin{align}
\left\{
    \begin{array}{lr}
    \mathbf{y}_{1} =\text{FFN}_{1}(\mathbf{y})=\text{ReLU}(\mathbf{y}\mathbf{W}_{1}+\mathbf{b}_{1}), & \\
    \mathbf{y}_{2}  =\text{FFN}_{2}(\mathbf{y}_{1})=\mathbf{y}_{1}\mathbf{W}_{2}+\mathbf{b}_{2}, &
    \end{array}
\right.
\end{align}
where $\mathbf{W}_{1}\in\mathbb{R}^{d\times d'}$ and $\mathbf{W}_{2}\in\mathbb{R}^{d'\times d}$ are weight matrices with $d'>d$, and $\mathbf{b}_{1}$ and $\mathbf{b}_{2}$ are bias vectors in FFN.

\subsection{Transformer Layer with Adaptive Sparsity}

We propose a general approach to integrate language-dependent sparsity into Transformer using latent variables. Borrowing the idea of layer selection \cite{li2020deep}, we use latent variables $\mathbf{z}$ to modulate component selection by languages.
The component can be an attention head, a block in FFN weight matrix, or an entire layer. Suppose that a sample $(x,y)$ has source text $x$ in language $l_{1}$ and target text $y$ in language $l_{2}$. The multilingual translation model has parameters $\boldsymbol{\theta}$ which are modulated by latent variables $\mathbf{z}$. 

\begin{align}
\label{eq:prob}
p(y|x,l_{1},l_{2},\boldsymbol{\theta}) = \mathbb{E}_{p(z|l_{1},l_{2},\boldsymbol{\theta})}[p(y|x,\textbf{z},l_{1},l_{2})].
\end{align}

Eq.~(\ref{eq:prob}) is intractable given too many choices in component selection. 
Latent variable $\mathbf{z}$ has the prior $p(\mathbf{z}|l_{1},l_{2})$. We parameterize it with $\boldsymbol{\mu}$ using Gumbel-Softmax trick \cite{DBLP:conf/iclr/JangGP17}, and estimates its posterior $q_{\boldsymbol{\mu}}(\mathbf{z}|l_{1},l_{2})$.

We derive the lower bound of Eq.~(\ref{eq:prob}) below with $\text{KL}(\cdot)$ as KL-divergence, and learn translation parameters $\boldsymbol{\theta}$ and selection parameters $\boldsymbol{\mu}$ to maximize the lower bound.
\begin{align}
\nonumber
\log p(y|x,l_{1},l_{2}) \geq  &\mathbb{E}_{q_{\boldsymbol{\mu}}(\mathbf{z}|l_{1},l_{2})}[\log p_{\boldsymbol{\theta}}(y|x,\mathbf{z},l_{1},l_{2})] \\
\label{eq:bound}
&-\text{KL}(q_{\boldsymbol{\mu}}(\mathbf{z}|l_{1},l_{2}) || p(\mathbf{z}|l_{1},l_{2})).
\end{align}

The derivation of inequality~(\ref{eq:bound}) is included in Appendix. We assume that each component $i$ is selected or discarded with equal probability by given languages, i.e., the prior $p(z_{i}=1|l_{1},l_{2})=p(z_{i}=0|l_{1},l_{2})=0.5$.



For simplicity, we denote the posterior of the $i$-th component, $q_{\mu_{i}}(z_{i}|l_{1}, l_{2})$ as score $s_{l,i}$, where $l$ is $l_{1}$ if the component $i$ is in encoder, and it is $l_{2}$ in decoder.
Score $s_{l,i}$ is also used to weigh the output of component $i$ in the forward pass of model computation.
In the following discussions, we distinguish the scores of different components with different notations. The score of an attention head is denoted as $\alpha$, the score of FFN block is $\beta$ and the score of a layer is $\gamma$. 

\noindent\textbf{Attention sparsity}. It has been revealed by previous studies that some attention heads are redundant in Transformer \cite{michel2019sixteen}. We propose to add sparsity to the attention module by masking partial heads. Suppose that language $l$ assigns a score $\alpha_{l,h}$ to the head $h$. The outputs $\mathbf{x}$ of attention heads are multiplied by their scores.  
\begin{align}
    \tilde{\mathbf{x}} = \alpha_{l,1}\mathbf{x}_{1} \oplus \cdots \oplus \alpha_{l,h}\mathbf{x}_{h} \oplus \cdots \oplus \alpha_{l,H}\mathbf{x}_{H},
\end{align}
where $l$ is the language which head scores are conditioned on. When the score is $0$, it is equivalent to setting dimensions corresponding to masked heads as $0$. Masked heads thus bring sparsity to resulting token vector $\tilde{\mathbf{x}}$.

\noindent\textbf{Feed-forward sparsity}. The feed-forward sub-layers transform token vectors by first increasing the hidden dimension and then converting it back to the original dimension. This suggests that the intermediate hidden state has redundancy due to the dimension increase from $d$ to $d'$. Hence we propose to sparsify the feed-forward matrices.

We divide $d'$ dimensions into $K$ blocks (sub-matrices), and each block has a width of $w=\frac{d'}{K}$ dimensions. A language $l$ selects a subset of blocks by assigning score $\beta_{l,k}$ to block $k$. Accordingly, the score matrix $\boldsymbol\beta_{l,k}$ for block $k$ is a $d\times w$ matrix with all elements as the score $\beta_{l,k}$. The score mask $\boldsymbol{\beta_{l}}\in\mathbb{R}^{d\times d'}$ assigned by language $l$ to matrix $W_{1}$ is $\boldsymbol{\beta}_{l}=\left[\boldsymbol{\beta}_{l,1}, \ldots, \boldsymbol{\beta}_{l,k}, \ldots, \boldsymbol{{\beta}}_{l,K} \right]$.

We apply the mask $\mathbf{\beta}_{l}$ to feed-forward matrices $\mathbf{W}_{1}$ and $\mathbf{W}_{2}$, and add sparsity by zeroing out their columns and rows respectively.
\begin{align}
\left\{
    \begin{array}{lr}
    \tilde{\mathbf{y}}_{1} =\text{FFN}_{1}(\mathbf{y})=\text{ReLU}(\mathbf{y}(\mathbf{W}_{1}\odot\boldsymbol{\beta}_{l})+\mathbf{b}_{1}), & \\
    \tilde{\mathbf{y}}_{2}  =\text{FFN}_{2}(\tilde{\mathbf{y}}_{1})=\tilde{\mathbf{y}}_{1}(\boldsymbol{\beta}_{l}^{T}\odot \mathbf{W}_{2})+\mathbf{b}_{2}, &
    \end{array}
\right.
\end{align}
where $\odot$ is element-wise multiplication.

\subsection{Adaptive Transformer}

So far we have discussed sparsity within a Transformer layer, and will continue towards sparsity across layers. We now introduce masks to the whole Transformer layer. Similar to \cite{li2020deep}, each language selects a subset of layers in the adaptive model. Suppose that language $l$ assigns score $\gamma_{l,r}$ to layer $r$ to indicate how likely it uses this layer. Suppose that the input (a sequence of token vectors) to layer $r$ is $\mathbf{u}_{r}$, and it is then processed by attention module ``$\text{Attn}$'' and feed-forward module ``$\text{FFN}$''. The output of layer $r$ , which is also the input to layer $r+1$, is denoted as $\mathbf{u}_{r+1}$.
\begin{align}
\left\{
    \begin{array}{lr}
    \mathbf{v} =\mathbf{u}_{r}+\gamma_{l,r}\cdot\text{Attn}(\text{LayerNorm}(\mathbf{u}_{r})), & \\
    \mathbf{u}_{r+1}  =\mathbf{v} + \gamma_{l,r}\cdot\text{FFN}(\text{LayerNorm}(\mathbf{v})), &
    \end{array}
\right.
\end{align}
where $\text{LayerNorm}(\cdot)$ is a normalization layer.

It is equivalent to a standard Transformer when $\gamma_{l,r}=1$. When the score $\gamma_{l,r}$ is $0$, layer $r$ is not selected and we have $\mathbf{u}_{r+1}=\mathbf{u}_{r}$. Accordingly, the model routes input $\mathbf{u}_{r}$ directly to layer $r+1$, skipping layer $r$.


\section{Training and Inference}


\subsection{Training Objective}
The commonly used training objective for machine translation is cross-entropy loss, which corresponds to the first part of the lower bound in ineqaulity~(\ref{eq:bound}). 
Besides the cross-entropy loss $L_{\text{cp}}$, our model adopts auxiliary losses to accommodate the adaptive and sparse architecture.

Suppose that we have $L$ language directions, and that the full multilingual model contains $D$ layers with $H$ attention heads and $K$ feed-forward blocks per layer. For simplicity of notation, we again denote the component score as $s$ in place of $\alpha$, $\beta$ and $\gamma$ used by different components. 

\noindent\textbf{Sparsity loss $L_{s}$}. The sparsity loss corresponds to the KL-divergence in the lower bound of inequality~(\ref{eq:bound}).
\begin{align}
    L_{s} = \sum\limits_{l=1}^{L}\sum\limits_{i=1}^{N}s_{l,i}(\log s_{l,i}-\log 0.5),
\end{align}
With the sparsity loss, the model leverages sparsity by assigning low scores to unimportant components.

\noindent\textbf{Disparity loss $L_{d}$}. Besides interference, too much parameter sharing among languages also leads to a waste of model capacity when some components are not used by any language at all. This motivates the disparity loss, which measures the similarity of module selection between languages.
\begin{align}
    L_{d} = \sum\limits_{l=1}^{L}\sum\limits_{l'=l+1}^{L}(\sum\limits_{i=1}^{N}s_{l,i}\cdot s_{l',i}),
\end{align}
where $N$ is the number of components that languages can choose.
Disparity loss is designed to encourage languages to choose different components to mitigate interference.

\noindent\textbf{Top-k loss $L_{t}$}.
For the sub-network activated for each language within our sparse model, the inference budget decides its number of layers $D' (D'<D)$, the number of attention heads $H' (H'<H)$ and feed-forward blocks $K' (K'<K)$ kept in each layer. With the learned scores $[s_{l,1}, \ldots, s_{l,N}]$ that language $l$ assigns to $N$ components, 
we select components with the highest scores in their category. For example, top $H'$ heads are selected among $H$ heads within a layer.
The selection is denoted by a binary vector $[m_{l,1}, \ldots, m_{l, N}]$, and $m_{l,i}$ is $1$ if the corresponding component is selected, and it is $0$ otherwise. We design the top-k loss $L_{t}$ to measure the difference between scores and binary masks.
\begin{align}
    L_{t} = \sum\limits_{l=1}^{L}\sum\limits_{i=1}^{N}(s_{l,i} - m_{l,i})^{2}.
\end{align}
By minimizing the top-k loss, each language selects exact $D'$ layers, $H'$ heads and $K'$ blocks as its activated sub-network to meet the inference efficiency requirements.

We leverage these losses in model training as will be discussed below.

\subsection{Training}

\begin{algorithm}[H]
\caption{Training Adaptive Sparse Model}
\label{algo}
\begin{algorithmic}[1]
\State{\textbf{Input}: Training data $\mathbf{D}_{\text{train}}$}, hyperparameters $c_{s}$, $c_{d}$ $c_{t}$, $\bar{T}$ and $T$
\State{\textbf{Output}: Multilingual model parameters $\boldsymbol{\theta}^{*}$, parameters $\boldsymbol{\mu}^{*}$ for attention head, FFN block and layer selection}
\For{$t=1, \ldots, \bar{T}$}
\State{Data batch: $(x, y) \leftarrow \text{Sample}(D_{\text{train}})$}
\State{Component scores: $\mathbf{s}_{t}\leftarrow \text{Gumbel-Softmax}(\boldsymbol{\mu}_{t})$}
\State{$L\leftarrow L_{\text{cp}}(x,y;\boldsymbol{\theta_{t}}, \mathbf{s}_{t})+c_{s}\cdot L_{s}(\mathbf{s}_{t})$}
\State{$\boldsymbol{\theta}_{t+1}, \boldsymbol{\mu}_{t+1}\leftarrow\text{update}(L; \boldsymbol{\theta_{t}},\boldsymbol{\mu}_{t})$}
\EndFor
\For{$t=\bar{T}+1, \ldots, T$} 
\State{Data batch: $(x, y) \leftarrow \text{Sample}(D_{\text{train}})$}
\State{Component scores: $\mathbf{s}_{t}\leftarrow\text{Gumbel-Softmax}(\boldsymbol{\mu}_{t})$}
\State{Binary masks: $\mathbf{m}_{t}\leftarrow \{1$ if $s_{t,i}$ is among top-k scores else $0$: $s_{t,i}\in\mathbf{s}_{t}$\}}
\State{Sparse scores: $\bar{\mathbf{s}}_{t}\leftarrow \mathbf{m}_{t}\odot\mathbf{s}_{t}$}
\State{$L\leftarrow L_{\text{cp}}(x,y;\boldsymbol{\theta}_{t}, \bar{\mathbf{s}}_{t})+c_{d}\cdot L_{d}(\bar{\mathbf{s}}_{t})+c_{t}\cdot L_{t}(\mathbf{s}_{t},\mathbf{m}_{t})$}
\State{$\boldsymbol{\theta}_{t+1}, \boldsymbol{\mu}_{t+1}\leftarrow\text{update}(L; \boldsymbol{\theta_{t}},\boldsymbol{\mu}_{t})$}
\EndFor
\end{algorithmic}
\end{algorithm}

\begin{table*}[htbp!]
\centering
\resizebox{1.0\textwidth}{!}{
\begin{tabular}{lcccccc}
\thickhline
\textbf{Model} & \multicolumn{3}{c}{\textbf{O2M}} & \multicolumn{3}{c}{\textbf{M2O}} \\ \hline
\textbf{Sparse model} & \textbf{\#Params (M)} & \textbf{Decode (tok/s)} & \textbf{BLEU} & \textbf{\#Params (M)} & \textbf{Decode (tok/s)} & \textbf{BLEU} \\ 
+ Attn+FFN+Layer & 31.5 (63.1) & 1439.8 & \textbf{19.8} & 31.5 (56.8) & 1402.5 & 22.4 \\
+ Attn only & 31.5 (31.5) & 1447.2 & 19.5 & 31.5 (31.5) & 1430.0 & \textbf{22.9} \\
+ FFN only & 31.5 (37.8) & 1401.3 & \textbf{19.8} & 31.5 (37.8) & 1360.0 & 22.8  \\ 
+ Layer only & 31.5 (50.5) & 1408.3 & 19.6 & 31.5 (44.2) & 1359.6 & 22.2 \\ \hline\hline
\textbf{Multi-Encoder} & - & - & - & 31.5 (119.0) & 1368.0 & 22.0 \\
\textbf{Multi-Decoder} & 31.5 (164.5) & 1421.3 & 19.5 & - & - & - \\
\textbf{Adapter} & 31.5 (50.9) & 1236.6 & 19.5 & 31.5 (50.9) & 1102.3 & 22.2 \\ \hline
\textbf{Transformer} & 31.5 (31.5) & 1412.4 & 19.3 & 31.5 (31.5) & 1381.9 & 22.0
\\\thickhline
\end{tabular}}
\caption{Average BLEU scores over $24$ translation directions on Public-24 data.  Besides the number of active parameters, total parameter sizes are included in the bracket. The inference efficiency is measured by the decoding speed, i.e., the number of decoded tokens per second.}
\label{tab:results_wmt}
\end{table*}

The model training is described in Algorithm \ref{algo}. We first train the model with both cross-entropy and sparsity loss. All model parameters are used by each language, and the model learns to score each component based on their impact on translation quality.
After $\bar{T}$ steps, the model starts component selection, i.e., only a specific number of components are used in model computation for a given language to meet the budget of inference cost. It is trained with cross-entropy loss, disparity loss and top-k loss. The weights on auxiliary losses are hyperparameters set as $c_{d}=0.02$, $c_{s}=0.1$, and $c_{t}=0.1$ in our experiments.

\subsection{Inference}
During inference, only selected components are activated given a language. The sub-network for inference consists of the selected $D'$ Transformer layers with $H'$ attention heads and $K'$ feed-forward blocks in each layer.

\section{Experiments}

We evaluate the proposed sparse models on multilingual translation including on one-to-many (O2M), many-to-one (M2O) and many-to-many (M2M) translation. O2M translation has source texts in one language and target texts in multiple languages. M2O translation has multiple source languages and only one target language, and M2M translation covers multiple source and target languages.

\subsection{Experimental Setup}

\noindent\textbf{Datasets}. Models are evaluated on two widely used multilingual translation datasets at different scales. More details of these datasets are included in Appendix. 

\begin{itemize}
    \item Public-24. This \textit{medium-scale} dataset contains parallel corpora between English and $24$ languages, collected from public sources such as WMT shared tasks \cite{liu2020multilingual}. It provides O2M and M2O translations.
    \item OPUS-100. This is a \textit{large-scale} multilingual translation dataset covering $100$ languages \cite{zhang2020improving}. It serves for M2M translations.
\end{itemize}  


\noindent\textbf{Baselines}. We include the following strong baselines which are commonly used in multilingual translation. 
\begin{itemize}
    \item Multilingual Transformer \cite{DBLP:conf/nips/VaswaniSPUJGKP17}. A single Transformer model for multilingual translation shares all parameters among languages.
    \item Multi-decoder Transformer \cite{sen2019multilingual,xiang2021multi}. Similar to the multilingual Transformer, it has an encoder-decoder architecture, but replaces the decoder with multiple decoders. It is used for one-to-many translation, and target languages are clustered into families based on their proximity \cite{lewis2009ethnologue}. One decoder is shared by target languages from the same family, and each family has an exclusive decoder.
    \item Multi-encoder Transformer. Similar to multi-decoder Transformer, it replaces a single encoder with multiple encoders. It is used for many-to-one translation, and each encoder corresponds to one family of source languages.
    \item Adapter based Transformer \cite{bapna2019simple}. Adapter layers are transplanted between adjacent layers of a trained multilingual Transformer. Each language pair is routed to its corresponding adapter layers which are finetuned on the same data with other Transformer parameters frozen. 
\end{itemize}

\begin{table*}[htbp!]
\centering
\resizebox{1.0\textwidth}{!}{
\begin{tabular}{lcccccccc}
\thickhline
\textbf{Model} & \textbf{\#Params (M)} & \multicolumn{2}{c}{\textbf{O2M}} & \multicolumn{2}{c}{\textbf{M2O}} &
\multicolumn{2}{c}{\textbf{Zero-shot}} \\ \hline
\textbf{Sparse model} & &  \textbf{Decode (tok/s)} & \textbf{BLEU} & \textbf{Decode (tok/s)} & \textbf{BLEU} & \textbf{Decode (tok/s)} & \textbf{BLEU} \\ 
+ Attn+FFN+Layer & 44.2 (138.7) & 1888.3 & \textbf{26.4} & 1938.1 & \textbf{31.5} & 1321.4 & \textbf{8.9} \\
+ Attn only & 44.2 (44.2) & 1919.4 & 24.6 & 2016.1 & 30.5 & 1405.1 & 4.0 \\
+ FFN only & 44.2 (69.4) & 1842.6 & 26.0 & 1966.7 & 31.2 & 1271.1 & 3.0 \\ 
+ Layer only & 44.2 (88.3) & 1860.5 & 25.6 & 1922.3 & 31.2 & 1184.4 & 4.9 \\ \hline\hline
\textbf{Adapter} & 44.6 (125.7) & 1606.5 & 26.1 & 1765.8 & 30.9 & - & - \\ \hline
\textbf{Transformer} & 44.2 (44.2) & 1858.4 & 24.3 & 1937.2 & 30.2 & 1206.1 & 2.7 \\
\thickhline
\end{tabular}}
\caption{Average BLEU scores over $94$ O2M and M2O translation directions along with zero-shot translation in $30$ directions on OPUS-100 dataset. We report the number of effective parameters and include the total parameter sizes in the bracket.}
\label{tab:results_opus}
\end{table*}

\noindent\textbf{Evaluation metrics}. As we consider both translation quality and model efficiency, two metrics are used for model evaluation. BLEU measures the translation quality by comparing the predicted and reference translations. As for the efficiency, we report the decoding speed, i.e., the number of decoded tokens per second (tok/s) when one GPU is used with a batch size of $4096$ tokens during inference. 

\subsection{Model and Training}
For multi-encoder (or multi-decoder), the number of encoders (or decoders) is the number of language families which are obtained using linguistic knowledge \cite{lewis2009ethnologue}. On Public-24 dataset, 24 languages are grouped into 8 families, and the grouping is included in Appendix.

In the adapter model, an adapter layer consists of two feed-forward sub-layers. 
The intermediate feed-forward dimension is set as $128$ in Public-24 experiments, and $32$ in OPUS-100 so that the adapter models have similar sizes as our sparse models for a fair comparison.

Our sparse model with layer sparsity starts with $6$ encoder layer and $12$ decoder layers for O2M translation, and $12$ encoder layers and $6$ decoders for M2O. It begins with $12$ encoder and decoder layers for M2M translation. During inference, only $6$ encoder and $6$ decoder layers are used. We note that the model with layer sparsity only is comparable to the results of latent depth model in \cite{li2020deep}.

As for the sparse model with attention sparsity, it has the same number of attention heads as Transformer baseline. 
On the Public-24 dataset, $3$ out of $4$ attention heads are selected in each Transformer layer. As for OPUS-100, $6$ out of $8$ heads are selected. Attention outputs corresponding to the unselected heads are masked with $0$'s. 

The model with feed-forward sparsity has a dimension of $2048$ on Public-24 data. 
The feed-forward matrix is divided into $8$ blocks, and $4$ blocks ($1024$ dimensions) are activated in each layer. For OPUS-100, the feed-forward sub-layer with a dimension of $4096$ is divided into $16$ blocks and $8$ blocks ($2048$ dimensions) are selected by each language. In sparse model training, hyperparameter $\bar{T}$ in Algo.~\ref{algo} is set as $8$k on Public-24 and $50$k on OPUS-100.

We report more training details and hyperparameters settings in Appendix. 

\subsection{Results}

To provide a good understanding of how sparsity in different parts of Transformer influences the multilingual performance, we experiment with sparsity in different components including attention sparsity (``Attn only''), feed-forward sparsity (``FFN only''), layer sparsity (``Layer only'') as well sparsity in all these components (``Attn+FFN+Layer''). 

We compare sparse models to dense baselines with the similar amount of active parameters during inference. We also report the total number of parameters including unused parameters which do not contribute to the computation cost but account for total memory usage of sparse models. We note that embeddings are excluded from parameter counting. We want to provide a consistent view of model capacity without being affected by embedding parameters varying with the vocabulary sizes across datasets.

\textbf{Public-24}. Table~\ref{tab:results_wmt} reports results on Public-24 dataset. 
In terms of BLEU score, the best models on O2M translation are sparse models with feed-forward sparsity and with all sparsity. They demonstrate an average $0.5$ BLEU gain over Transformer in $24$ language directions. As for M2O translation, the best performance is achieved by the sparse model with attention sparsity. Its improvement over Transformer is $0.9$ BLEU. Adapter model has comparable performance to multi-encoder and multi-decoder model, and outperforms Transformer by $0.2$ in both O2M and M2O translations.

As for decoding speed, sparse models with only feed-forward sparsity and with only layer sparsity are as efficient as Transformer baseline, as their activated sub-networks have the same architecture as Transformer. The sparse model with attention sparsity improves efficiency over Transformer in that fewer attention heads are activated during inference.
Multi-encoder and multi-decoder models also have comparable decoding speed as Transformer. Adapter model is slower in inference in comparison with other models due to extra computation costs brought by $12$ adapter layers.

\textbf{OPUS-100}. The results on OPUS-100 are shown in Table~\ref{tab:results_opus}. We did not include multi-encoder or multi-decoder since they are too large for efficient training.  The models are trained on M2M translations, and we report the average BLEU scores on the test set in $94$ one-to-many directions, $94$ many-to-one and $30$ zero-shot directions. Adapter model trains adapter layers for each language pair, and cannot be applied to zero-shot directions without training data.

It can be seen from Table~\ref{tab:results_opus} that the sparse model achieves the best BLEU by combining all types of sparsity in attention, feed-forward and layer. In comparison with Transformer, it achieves an average of $+2.1$, $+1.3$ and $+6.2$ BLEU in O2M, M2O and zero-shot translations respectively. It also outperforms Adapter by +$0.3$ and +$0.6$ BLEU in O2M and M2O translations respectively. 

As for inference efficiency, we again observe that sparse model with attention sparsity is fastest in decoding. It is worth mentioning that attention sparsity yields better BLEU with higher efficiency when compared with Transformer baseline. Sparse models with feed-forward sparsity or layer sparsity demonstrate comparable inference efficiency to Transformer, and achieve faster inference than Adapter.



\section{Discussion}
\subsection{Ablation Study}

\textbf{Auxiliary losses}. Three auxiliary losses are leveraged in training. We perform ablation studies to measure the impact of each loss on Public-24 O2M translation, using the model with FFN sparsity.
In Table~\ref{tab:aux}, we report BLEU of each model when one auxiliary loss is removed from their training objective. 

\begin{table}[htbp!]
\centering
\begin{tabular}{l|c}
\thickhline
Sparse FFN Model & BLEU \\ \hline
All losses & 19.8 \\
- Sparsity loss & 19.7 \\
- Top-k loss & 19.3 \\
- Disparity loss & 19.5 \\ \thickhline
\end{tabular}
\caption{Ablation study of auxiliary losses on models with FFN sparsity in Public-24 one-to-many translation.}
\label{tab:aux}
\end{table}

The model trained without top-k loss shows the largest drop of $0.5$ BLEU. Top-k loss trains the model so that the learned architecture satisfies the sparsity requirements such as the number of activated layers and feed-forward dimensions. Without disparity loss, the model loses $0.3$ BLEU as it is not punished for sharing parameters among languages. The performance drop without disparity loss results from the language interference in shared parameters. A drop of $0.1$ BLEU is observed without sparsity loss. We note that the top-k loss compensates for the exclusion of sparsity loss, as it also encourages sparse architecture.

\noindent\textbf{Sparsity types}. As shown in Table~ \ref{tab:results_wmt} and \ref{tab:results_opus}, sparsity in different model components affects translation quality and efficiency differently.
We now compare and analyze these sparsity types from the results in rows of ``Attn only'', ``FFN only'' and ``Layer only''. The sparsity in each component alone improves BLEU scores in comparison with multilingual Transformer. This suggests that the language interference exists in all these components.

With O2M translation on Public-24 data, FFN sparsity yields better BLEU than layer sparsity with fewer parameters. The improvement of attention sparsity is relatively smaller compared with layer and FFN sparsity for O2M translation. 
As for M2O translation, we observe that attention sparsity outperforms other types of sparsity on Public-24 dataset.
In this medium-scale dataset with $24$ languages, positive transfer brought by parameter sharing is more obvious than the negative interference in both O2M and O2M translations. This is suggested by the fact that multi-encoder and layer sparsity are beaten by other sparse models. Both of them are coarse-grained approaches of parameter sharing. It reduces many shared parameters when two languages are routed to independent encoders or layers. However, when sparsity is applied to attention heads or feed-forward blocks, the model could benefit from knowledge transfer via partially shared parameters within a component.

When it comes to large-scale OPUS-100, the model combining all types of sparsity demonstrate the best BLEU in O2M, M2O and zero-shot translations. A possible explanation is that negative inference is dominant in a multilingual model supporting a large number of languages. Mitigation of interference by reducing parameter sharing is more effective in improving translations. We note that larger BLEU gains are observed in O2M, M2O and zero-shot translations as the data scale grows larger. It justifies the effectiveness of our parameter sharing approach when scaled to numerous languages. We include additional analysis in Appendix. 

\subsection{Sparsity Pattern}

\begin{figure}[htbp!]
\centering
\includegraphics[width=0.4\textwidth]{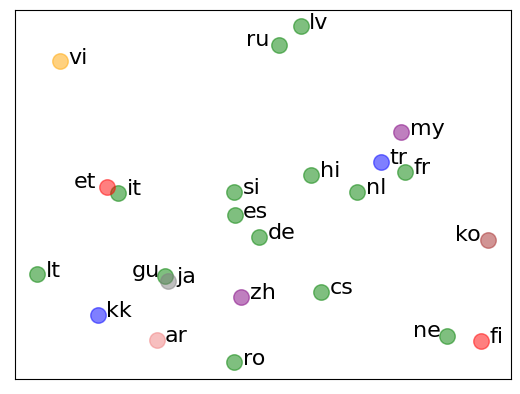}
\caption{Visualization of languages based on their model component selection. Language families are differentiated with different colors (Pink: Arabic, blue: Turkic, orange: Austroasiatic, green: Indo-European, red: Uralic, purple: Sino-Tibetan, brown: Korean, grey: Japonic).}
\label{fig:group}
\end{figure}
To gain an insight into the language-conditioned sparsity in our sparse models, we now analyze the component selection by different languages. Take the sparse model trained on O2M translation of Public-24 dataset as an example, which integrates sparsity into layers, attention and FFN sub-layers. For a given language, the model assign its score to each component. We represent this language with a vector of all component scores, and visualize these language vectors in Fig.~\ref{fig:group} with principal component analysis (PCA). The closeness of languages reflects the similarity of their component selection patterns.

It is observed that languages in Indo-European family form multiple clusters. One such cluster consists of Russian (ru) and Latvian (lv). Some languages such as Vietnamese (vi) and Korean (ko) stay away from languages in other families. The parameter sharing is shown correlated with language proximity to some extent. 

Interestingly, many high-resource languages including Czech (cs), Chinese (zh), German (de) and Russian (ru) do not have much parameter sharing with low-resource languages. As can be seen in Fig.~\ref{fig:group}, they do not have low-resource as close neighbors. This suggests that parameter sharing is also related to the resource sizes of languages. It resonates with previous findings that low-resource degrades the performance of high-resource languages in multilingual models \cite{wang2020negative}. 
Exceptions are French (fr) with the low-resource neighbor Turkish (tr) and Spanish (es) with low-resource neighbor Sinhala (si).

\section{Conclusion}

In this study, we propose adaptive language-conditioned sparsity for multilingual translation. Our fine-grained parameter sharing strategy is a general approach to integrate and compare sparsity at different scales, from attention heads, FFN blocks to Transformer layers. 
With extensive experiments across multiple datasets, our sparse models demonstrate consistent performance gains over strong baselines including multilingual Transformer without increasing the inference cost.
One limitations of this work is the lack of analysis about how model quality and efficiency change with sparsity, which would be addressed in future work.


\bibliographystyle{acl_natbib}
\bibliography{acl2021}

~\newpage

\appendix

\section{Sparse Model}
\label{app:model}

\noindent\textbf{Derivation of lower bound}. We now provide the evidence lower bound of Eq.~(\ref{eq:prob}) below.

\begin{align}
\nonumber
&\mathbb{E}_{q_{\boldsymbol{\mu}}(\mathbf{z}|l_{1},l_{2})}[\log p_{\boldsymbol{\theta}}(y|x,\mathbf{z},l_{1},l_{2})] \\
\nonumber
&-\text{KL}(q_{\boldsymbol{\mu}}(\mathbf{z}|l_{1},l_{2}) || p(\mathbf{z}|l_{1},l_{2})) \\
\nonumber
&=\mathbb{E}_{q_{\boldsymbol{\mu}}(\mathbf{z}|l_{1},l_{2})}[\log p_{\boldsymbol{\theta}}(y|x,\mathbf{z},l_{1},l_{2})] \\
\label{eq:kl}
&-\mathbb{E}_{q_{\boldsymbol{\mu}}(\mathbf{z}|l_{1},l_{2})}[\log q_{\boldsymbol{\mu}}(\mathbf{z}|l_{1},l_{2})- \log p(\mathbf{z}|l_{1},l_{2})] \\
\nonumber
&=\mathbb{E}_{q_{\boldsymbol{\mu}}(\mathbf{z}|l_{1},l_{2})}\left[\log p_{\boldsymbol{\theta}}(y|x,\mathbf{z},l_{1},l_{2}) + \log p(\mathbf{z}|l_{1},l_{2})\right. \\
\nonumber
& \left. - \log q_{\boldsymbol{\mu}}(\mathbf{z}|l_{1},l_{2})\right] \\
\label{eq:jensen}
&\leq \log\mathbb{E}_{q_{\boldsymbol{\mu}}(\mathbf{z}|l_{1},l_{2})}[\frac{p_{\boldsymbol{\theta}}(y|x,\mathbf{z},l_{1},l_{2})\cdot p(\mathbf{z}|l_{1},l_{2})}{q_{\boldsymbol{\mu}}(\mathbf{z}|l_{1},l_{2})}] \\
\nonumber
&\leq \log\left(\sum\limits_{\mathbf{z}} p_{\boldsymbol{\theta}}(y|x,\mathbf{z},l_{1},l_{2})\cdot p(\mathbf{z}|l_{1},l_{2})\right) \\
&\leq \log p(y|x, l_{1}, l_{2}).
\end{align}

We note that the equation (\ref{eq:kl}) is based on the definition of KL-divergence. As for the inequality (\ref{eq:jensen}), it is based on Jensen's inequality. Hence we have proved the inequality~(\ref{eq:bound}):
\begin{align}
\nonumber
\log p(y|x,l_{1},l_{2}) \geq  &\mathbb{E}_{q_{\boldsymbol{\mu}}(\mathbf{z}|l_{1},l_{2})}[\log p_{\boldsymbol{\theta}}(y|x,\mathbf{z},l_{1},l_{2})] \\
\nonumber
&-\text{KL}(q_{\boldsymbol{\mu}}(\mathbf{z}|l_{1},l_{2}) || p(\mathbf{z}|l_{1},l_{2})).
\end{align}

\noindent\textbf{Model complexity}. The component selection in the proposed sparse models is lightweight in that only a small number of extra parameters are introduced to the model. Suppose that the sparse model has a depth of $D$ layers, each layer has $H$ attention heads, and each FFN module is divided into $K$ blocks. Given $L$ languages, the number of parameters for head selection is $D\times H\times L$, and the number of parameters for FFN block selection is $D\times K\times L$, and the parameter size for layer selection is $D\times L$. As we can see, the extra parameters introduced to the adaptive sparse model is much fewer compared with the model size.

\section{Datasets}
\label{app:dataset}

\begin{table*}[ht]
\centering
\begin{tabular}{ccccccccc}
\thickhline\\
Language & Code & Size & Source & Language & Code & Size & Source \\ \hline
Gujarati & gu & 10k & WMT19 & Kazakh & kk & 91k & WMT19 \\
Vietnamese & vi & 133k & IWSLT15 & Turkish & tr & 207k & WMT17 \\
Japanese & ja & 223k & IWSLT17 & Korean & ko & 230k & IWSLT17 \\
Dutch & nl & 237k & IWSLT17 & Arabic & ar & 250k & IWSLT17 \\
Italian & it & 250k & IWSLT17 & Burmese & my & 259k & WAT19\\
Nepali & ne & 564k & FLoRes & Romanian & ro & 608k & WMT16 \\
Sinhala & si & 647k & FLoRes & Hindi & hi & 1.56M & ITTB \\
Estonian & et & 1.94M & WMT18 & Lithuanian & lt & 2.11M & WMT19 \\
Finnish & fi & 2.66M & WMT17 & Latvian & lv & 4.50M & WMT17 \\
Czech & Cs & 11M & WMT & Spanish & es & 15M & WMT \\
Chinese & zh & 25M & WMT & German & de & 28M & WMT \\
Russian & ru & 29M & WMT & French & fr & 41M & WMT \\ \thickhline
\end{tabular}
\caption{Data statistics of public-24 dataset.}
\label{tab:wmt_data}
\end{table*}

The Public-24 dataset is recently collected by \cite{liu2020multilingual} from multiple public sources as shown in Table \ref{tab:wmt_data}. The sources are WMT shared tasks, IWSLT competition, WAT, FloRes and ITTB.

The OPUS corpus comes from multiple sources including movie subtitles, GNOME documentation and Bible. Following the data sampling process in \cite{zhang2020improving}, we prepare the OPUS dataset with up to $1$M sentence pairs per language pair for training, $2$k for validation and $2$k for testing. A total of $55$M sentence pairs are included in the OPUS dataset.

\section{Experiments}
\label{app:experiments}

\subsection{Empirical Setup}

For multi-encoder and multi-decoder models, languages are grouped into families based on the their proximity. Multi-encoder is used for many-to-one translation, sharing an encoder for source languages in the same family. Similarly for multi-decoder Transformer, one decoder is assigned to a family of target languages.

Public-24 data contains eight families: (1) Arabic; (2) Kazakh and Turkish; (3) Vietnamese; (4) Czech, German, Spanish, French, Gujarati, Hindi, Italian, Lithuanian, Latvian, Nepali, Dutch, Romanian, Russian and Sinhala; (5) Estonian and Estonian; (6) Chinese and Burmese; (7) Korean; (8) Japanese.

\textbf{Hyperparameters}. We sample translation data in different languages with a temperature $\tau$ of $5.0$ due to the data imbalance during training. Source and target vocabularies are learned with sentence-piece model and prepared for each dataset. Public-24 dataset has $250$k tokens and OPUS has $64$k tokens. We note that vocabulary size affects the model size since it decides the number of embedding parameters in both encoder and decoder.

The models in our experiments are built upon Transformer architecture. They have $6$ encoder layers and $6$ decoder layers, following the setting in \cite{wang2020balancing}. The embedding dimension for both encoder and decoder is set as $512$. On Public-24 data, the number of attention heads is set as $4$ and the feed-forward dimension is $1024$.  For OPUS-100, the model has $8$ attention heads and feed-forward dimension of $2048$. The training batch size is set as $150$k tokens on all datasets.
As for decoding, the beam size is $5$ and length penalty is $1.0$ for TED8 and Public-24 data. The beam size is $4$ and length penalty is $0.6$ for OPUS-100.

On Public-24 dataset, we have a dropout probability of $0.1$, and a learning rate of $0.0007$. Models on Public-24 data are trained for $100$k steps. The models on OPUS data has a dropout of $0.1$, and are trained for $500$k steps with a learning rate of $0.0015$.

\subsection{Result Analysis}
\label{app:discussion}


\noindent\textbf{Resource size affects performance}.  The language interference is reflected by the performance drop in high-resource languages when trained together with low-resource languages \cite{conneau2019unsupervised}. We analyze how model performance varies with the resource size of languages in Table~\ref{tab:resource_results}. Following \cite{liu2020multilingual}, we divide languages in Public-24 dataset based on their data sizes: $13$ low-resource languages with fewer than 1M parallel sentences, $6$ high-resource languages with more than 10M sentences, and the remaining $5$ medium resource languages. The sparse models in Table ~\ref{tab:resource_results} are the model with feed-forward sparsity for O2M translation and the model with attention sparsity for M2O translation.

\begin{table}[htbp!]
\begin{tabular}{c|l|ccc}
\thickhline
 & Model & High & Med & Low \\ \hline
\multirow{4}{*}{O2M} & Sparse Model &   \textbf{26.2} & \textbf{13.3} & \textbf{19.3}  \\
 & Multi-Enc/Dec &  25.9 & 13.0 & 19.0  \\
 & Adapter & 25.5 & 13.0 & 19.2  \\
 & Transformer & 25.2 & 12.8 & 19.0  \\ \thickhline
\multirow{4}{*}{M2O} & Sparse Model & \textbf{29.9} & \textbf{17.0} & \textbf{22.7}  \\
 & Multi-Enc/Dec &  29.0 & 16.8 & 21.6  \\
 & Adapter & 29.2 & 16.7 & 22.0  \\
 & Transformer & 28.6 & 16.3 & 21.9 \\
\thickhline
\end{tabular}
\caption{BLEU scores on high, medium and low-resource languages in Public-24 dataset.}
\label{tab:resource_results}
\end{table}

Sparse models consistently improve the translation quality of all categories of languages compared against other baselines. This resonates with a recent finding that interference impacts not only high-resource but also low-resource languages \cite{wang2020negative}. Compared with Transformer baseline, all other models demonstrate similar trends of gains over different resources: the gain on high-resource $>$ medium-resource $>=$ low-resource. In particular, the sparse models achieve BLEU gains of $1.0$ and $1.3$ on high-resource languages in O2M and M2O translations respectively.

\end{document}